\pdfoutput=1

\documentclass[11pt]{article}

\usepackage{acl}

\usepackage{times}
\usepackage{latexsym}

\usepackage[T1]{fontenc}

\usepackage[utf8]{inputenc}

\usepackage{microtype}
\usepackage{inconsolata}
\usepackage{subfigure}
\usepackage{graphicx} 
\usepackage{multirow}
\usepackage{float}
\usepackage{amsmath}
\usepackage{microtype}
\usepackage{booktabs}

\usepackage{pifont}
\usepackage{xspace}
\usepackage{stfloats}
\usepackage{hyperref}

\usepackage{xcolor}

\newcommand{\cmark}{\ding{51}}
\newcommand{\xmark}{\ding{55}}
\newcommand{\DABS}{Disordered-DABS\xspace}
\newcommand{\DCNNDM}{D-CnnDM\xspace}
\newcommand{\DWIKIHOW}{D-WikiHow\xspace}
\newcommand{\AspDiff}{\#AbsAspDiff}
\newcommand{\Cluster}{\textit{Cluster}}
\newcommand{\Keyword}{\textit{Keyword}}
\newcommand{\Prompting}{\textit{Prompting}}

\title{\DABS{}: A Benchmark for Dynamic Aspect-Based Summarization in Disordered Texts}

\author{Xiaobo Guo \and Soroush Vosoughi\\
        Department of Computer Science \\ Dartmouth College \\
    Hanover, New Hampshire\\
    \{xiaobo.guo.gr, soroush.vosoughi\}@dartmouth.edu
    }
  
\begin{document}
\maketitle

\begin{abstract}

    Aspect-based summarization has seen significant advancements, especially in structured text. Yet, summarizing disordered, large-scale texts, like those found in social media and customer feedback, remains a significant challenge. Current research largely targets predefined aspects within structured texts, neglecting the complexities of dynamic and disordered environments. Addressing this gap, we introduce \DABS{}, a novel benchmark for dynamic aspect-based summarization tailored to unstructured text. Developed by adapting existing datasets for cost-efficiency and scalability, our comprehensive experiments and detailed human evaluations reveal that \DABS{} poses unique challenges to contemporary summarization models, including state-of-the-art language models such as GPT-3.5.
\end{abstract}

\section{Introduction}
    The exponential growth of digital content has significantly increased the importance of automated text summarization methods. These methods are crucial for distilling salient content from large volumes of text and efficiently addressing diverse information needs. Query-focused summarization (QFS)~\cite{wang2022squality,zhong2021qmsum} and aspect-based summarization (ABS)~\cite{hayashi2021wikiasp,ahuja2022aspectnews} have emerged as prominent approaches for creating focused summaries based on specific aspects or queries. This advancement has led to the development of several specialized datasets and benchmarks.

    While QFS and ABS are adept at generating summaries targeted at specific aspects, they typically depend on predefined aspects for querying or model fine-tuning. Dynamic Aspect-Based Summarization (DABS) was introduced to overcome this limitation. DABS identifies aspects dynamically from the content of source articles, transcending the restrictions of fixed aspect definitions.

    \begin{figure}[!hbt]
    \centering
    \includegraphics[width=.95\columnwidth]{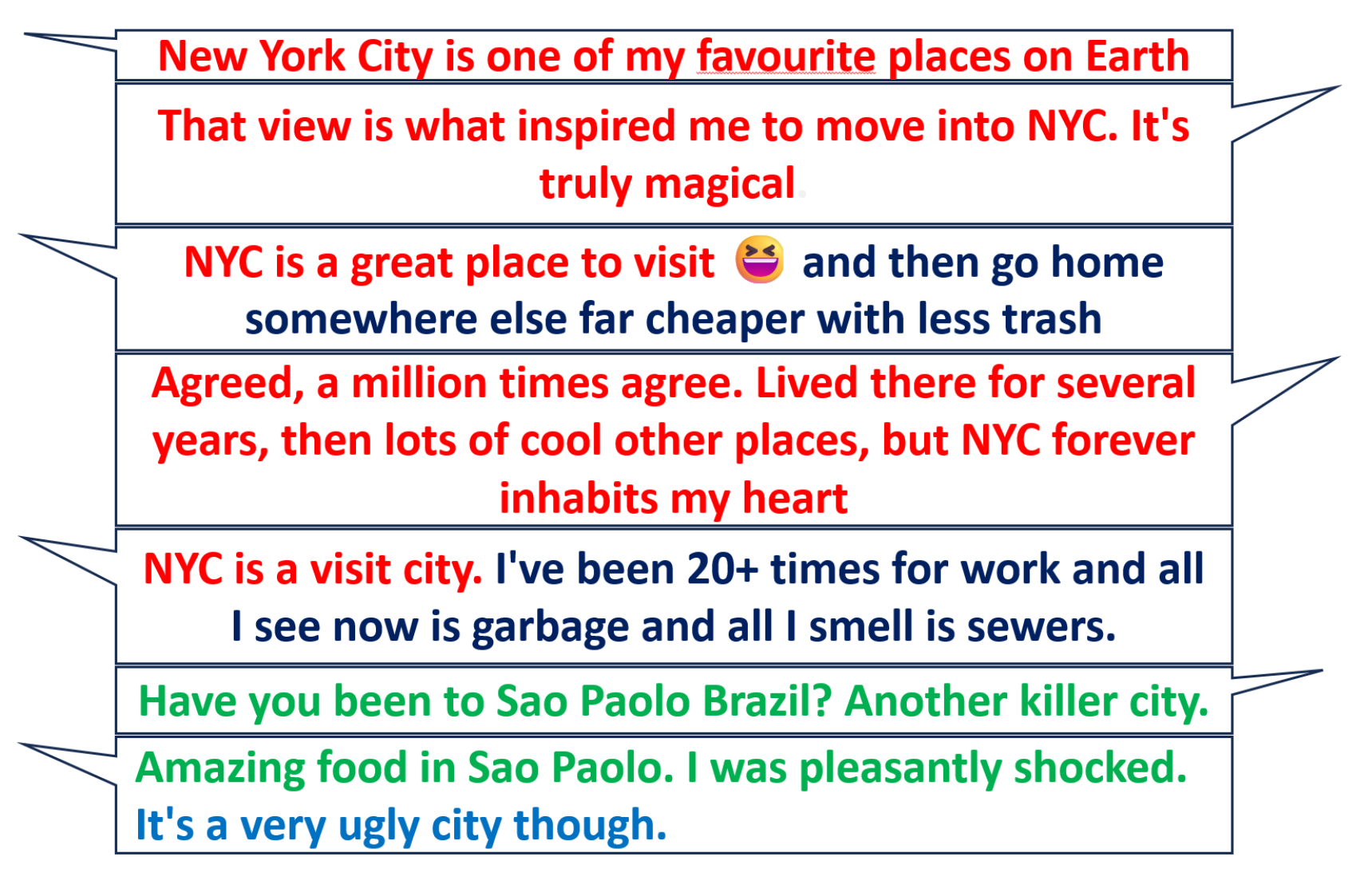}
    \caption{An example from Reddit showcasing a multi-user discussion. Each participant's input is color-coded to distinguish the varied aspects they discuss, emphasizing the range of perspectives and topics within the conversation. This visual differentiation serves to highlight the diversity inherent in such online discussions.}\label{fig:data_sample}
    \end{figure}

    Despite the innovations in DABS, a significant challenge remains, particularly in real-world applications: the blending of sentences from varied sources within a text. This complication is frequently observed in scenarios like summarizing community opinions, analyzing user feedback, or collating medical records from multiple sources. Such situations usually involve quick changes in topics or aspects, leading to a notable inconsistency in coherence among texts from different origins. Figure~\ref{fig:data_sample} illustrates a discussion from Reddit involving multiple users. The dialogue initially focuses on the appeal of New York City, transitions to its drawbacks for long-term residence, and eventually shifts to a conversation about Sao Paulo. This example underscores the dynamic nature of topics in online discussions, highlighting the complexity of summarizing such disordered texts \footnote{The full discussion can be seen on \href{https://www.reddit.com/r/travel/comments/x90kc4/new_york_city_is_one_of_my_favourite_places_on/}{Reddit}}. This disorder not only makes the summarization process more complex but also requires strategies adept at seamlessly merging disparate aspectual data, adding another layer of complexity to the task.

    Our paper addresses these challenges by presenting a novel and efficient method to modify existing summarization datasets, making them suitable for dynamic aspect-based summarization in disordered texts. We use this method to introduce the \DABS{} dataset\footnote{The data is available on \href{https://github.com/xiaobo-guo/Disordered-DABS}{Github}}, specifically curated for this demanding summarization task. Comprising 255,286 aspect-oriented summaries, this dataset is segmented into training, validation, and testing partitions. Its robustness is validated through comprehensive human evaluations. Our examination of various baseline models highlights the complexity of this task, revealing that even advanced large language models like GPT-3.5 face significant challenges in dealing with the nuances of \DABS{}. These observations underscore the intricacy inherent in this form of summarization, highlighting the necessity for more specialized datasets tailored to its unique requirements.

\section{Related Work}
    QFS has long been a fundamental task in the field of text summarization, focusing on generating summaries tailored to specific queries~\cite{dang2005overview}. It targets extracting precise information relevant to a wide range of queries, reflecting the diverse informational needs in various contexts. Key datasets in QFS, such as those developed by \citet{xu2021generating}, have been instrumental in addressing these needs, employing queries like ``steroid use among female athletes'' to explore detailed aspects such as ``trends'', ``side effects'', and ``consequences''.

    ABS emerged as an extension, initially focusing on customer review analysis~\cite{hu2004mining,lu2009rated}. ABS expands the scope of summarization to a broader range of domains, including news articles and encyclopedia content, as evidenced by works in ABS tasks~\cite{frermann2019inducing, ahuja2022aspectnews} and dataset creation~\cite{hayashi2021wikiasp}. This shift from query-specific summaries in QFS to aspect-oriented summaries in ABS demonstrates the evolution of summarization techniques in response to varying content and context requirements.

    Further advancing this evolution is DABS, which diverges from traditional ABS by not limiting aspects to predefined categories. Innovations in DABS, such as AnyAspect~\cite{tan2020summarizing} and ENTSUMv2~\cite{mehra2023entsumv2}, have focused on utilizing entities as aspects, with AnyAspect summarizing sentences around named entities and ENTSUMv2 employing human-annotated, entity-specific summaries. Other datasets like OASUM~\cite{yang-etal-2023-oasum} and OPENASP~\cite{amar2023openasp} explore conceptual labels or sub-topics as aspects, with OASUM leveraging Wikipedia subtitles and OPENASP adopting a multi-document DABS settings. A comparative overview of these datasets is presented in \autoref{tab: related_work_comparision}.

    \begin{table*}[!hbt]
\begin{tabular}{llclr}
\toprule
                \textbf{Dataset} & \textbf{Domain}  & \textbf{Disordered} & \textbf{Collection}  & \textbf{\# Instances} \\ \midrule
                AnyAspect {\small \citep{tan2020summarizing}}            & News                   & \xmark{}             & automatic                      & 312,085             \\
                              OASUM {\small \citep{yang-etal-2023-oasum}}            & Wikipedia                  & \xmark{}             & automatic                      & 3,747,569             \\ 
                              OpenASP{\small \citep{amar2023openasp}} & News      & \xmark{}    & manual            & 1,310       \\ 
                              ENTSUMV2{\small \citep{maddela2022entsum}} &News  &\xmark{} &automatic  &2,788 \\
                              \DABS{} (Ours) &News \& Wikipedia  &\cmark{} &automatic &255,286 \\
                              \bottomrule
\end{tabular}

\caption{Prominent datasets for dynamic aspect-based summarization.  ``\# Instances'' is the number of data instances in the dataset. ``Disordered'' means the sentences are shuffled in the source article}\label{tab: related_work_comparision}
\end{table*}

    However, a notable limitation of existing datasets in both ABS and DABS is their primary focus on well-organized and coherent texts, such as a news article. This focus is less relevant to platforms such as social media, where content often features multiple authors and rapid shifts in context. The inherent coherence in these texts could potentially overstate the performance of ABS and DABS models, as they may depend on sentence transitions to discern changes in topics or aspects. This coherence within texts of the available datasets might inadvertently inflate the performance of ABS and DABS models, as they might rely on transitions in contiguous sentences to infer changes in topics or aspects, whereas, in a dataset from multiple sources, this transition may detect the change of author and not aspect. Therefore, we propose the \DABS{} in which the aspects are dynamic and the texts are disordered.

\section{Dynamic Aspect-Based Summarization}
        Despite substantial efforts in QFS and ABS, the definition of ``aspect'' remains ambiguous. It ranges from different topics~\cite{frermann2019inducing,maddela2022entsum,tan2020summarizing} to specific elements within a topic~\cite{hayashi2021wikiasp,ahuja2022aspectnews,meng2021bringing}. Some studies use concise phrases for aspects~\cite{amar2023openasp,angelidis2021extractive}, while others allow longer sentences for complex information needs~\cite{zhong2022unsupervised,zhong2021qmsum}. Given this variability in aspect definition, our \DABS{} incorporates two distinct sub-datasets: one emphasizing aspect diversity across varied topics, and another focusing on different aspects of a single event, providing a comprehensive benchmark for these divergent aspect interpretations.

        Building on previous DABS research~\cite{tan2020summarizing, yang-etal-2023-oasum}, our task is defined as generating multiple aspect-based summaries from a document \(D\) containing \(n\) topics or aspects. Unlike traditional baselines that pair a document with a specific aspect for aspect-based summary generation~\cite{amar2023openasp}, our task requires the model to autonomously generate multiple summaries without predefined aspects. Additionally, to challenge the model's ability to handle non-coherent texts, we shuffle the sentences within \(D\), following \citet{frermann2019inducing}, disrupting the natural flow and requiring the model to cluster information from the entire document.

    \subsection{Automatic Evaluation Metrics}\label{sct: automatic evaluation metrics}
  
Automatic evaluation metrics are vital for the efficient development and evaluation of models for the ABS and DABS tasks. Our evaluation focuses on two primary factors: the capability to identify aspect-based information and the overall quality of the generated summaries. We measure the aspect identification accuracy using the absolute difference in aspects between the reference and generated summaries (\(\text{\AspDiff{}}\)). For assessing summary quality, we employ the standard summary quality metrics Rouge-1/2/L~\cite{lin2004rouge} and BERTScore~\cite{zhang2019bertscore}.

        To compare reference and generated summaries effectively, we pad the summaries to equalize the number of aspects. This approach penalizes models that either miss aspects or generate excessive ones. The optimal pairing of reference and generated summaries (\(\langle\hat{S}, S\rangle\)) is determined by maximizing the mean performance across all metrics, as defined by:
        \begin{equation}
        \langle\hat{S}, S\rangle = \text{argmax}\left(\sum_{\hat{s} \in \hat{S}, s \in S}\left(\sum_{m \in M} m(\widehat{s},{s})\right)\right)
        \end{equation}
        Here, \(S\) and \(\hat{S}\) denote the reference and generated summaries, respectively, and \(M\) represents the set of evaluation metrics.
    
    \subsection{Human Evaluation Metrics}\label{sct: human evaluation metrics}
        Human evaluation metrics offer a reference-free perspective, assessing the quality of the dataset and baselines using the source article. Evaluations focus on two aspects: overall summarization quality and aspect quality.
        
        Following \citet{fabbri2021summeval}, we assess summaries using four criteria: (1) \textit{Coherence}, evaluating the overall coherence of the summary; (2) \textit{Consistency}, evaluating factual alignment of the summary with the source; (3) \textit{Relevance}, focusing on the inclusion of key content from the source; and (4) \textit{Fluency}, examining the linguistic quality of the summary. 
        
        Furthermore, in the context of DABS, an additional metric, \textit{Aspect-Quality}, is introduced. This metric evaluates the distinctiveness and focus of aspect-based summaries, ensuring that each aspect is clearly represented and remains the central focus within its respective summary~\cite{angelidis2021extractive,amplayo2021aspect}.

\section{Datasets}\label{sct: datasets}
    \begin{table*}[hbt]
    \centering
    \setlength\tabcolsep{4pt}
    \begin{tabular}{llllllll}
    \toprule
    Dataset   & Domain       & \#Train & \#Valid & \#Test & Avg.Arc. Len       & Avg.Sum. Len & Avg.\# Asp \\ \midrule
    \DCNNDM{}   & News         & 47,920  & 2,185   & 1,917  & 4,143 (233)     & 311 (172) & 6.00 (3.15)   \\
    \DWIKIHOW{} & Encyclopedia & 142,284 & 20,327  & 40,653 & 452 (482)       & 41 (34) & 6.23 (4.70)   \\ \bottomrule
    \end{tabular}
    \caption{Statistics of the datasets. ``Avg.Arc.Len'', ``Avg.Sum.Len'', and ``Avg.\#Asp'' denote the average length of articles, summaries, and the average number of aspects per sample. Standard deviations are provided in brackets.
    }\label{tab: dataset_statistics}
\end{table*}

    The \DABS{} dataset comprises two subsets: \textbf{\DCNNDM{}} and \textbf{\DWIKIHOW{}}, each adapted from existing summarization datasets. Instead of constructing these subsets from the ground up, we opted for an automatic conversion approach, resulting in large-scale samples that are well-suited for fine-tuning purposes.

    \subsection{Dataset Creation}
         Each sub-dataset is created using distinct methods to generate dynamic aspect-based article-summary pairs from original article-summary pairs.

         \textbf{\DCNNDM{}} derivedfrom the CNN/DailyMail dataset~\cite{see2017get}, following the aggregation approach of \citet{frermann2019inducing}. Each sample in \DCNNDM{} consists of up to 11 randomly selected instances from CNN/DailyMail, treating each instance as a distinct aspect, thereby providing coarse-grained aspects.

        \textbf{\DWIKIHOW{}} modifies the WikiHow dataset~\cite{koupaee2018wikihow}, where each article is composed of multiple paragraphs, each introduced by a bold summary sentence. In \DWIKIHOW{}, each summary sentence is treated as an individual aspect.

    After aggregation, sentences within source articles are intentionally shuffled to produce disordered texts. This design aims to challenge models to cluster sentences by their aspects and generate precise aspect-based summaries. Such an approach not only tests the models' proficiency with ordered texts but also enhances their applicability to scenarios involving disordered texts.

        An overview of the datasets is provided in \autoref{tab: dataset_statistics}, highlighting the significant variability in article lengths, summary lengths, and aspect numbers. More detailed distribution statistics are available in Appendix~\ref{sct: Dataset-appendix}. The samples of the dataset are shown in Appendix~\ref{sct: DABS-samples}

    \subsection{Dataset Quality}

        We conducted a human evaluation of the datasets, adhering to the methodology described in Section~\ref{sct: human evaluation metrics}. For this assessment, we randomly selected thirty samples from each of \DCNNDM{} and \DWIKIHOW{}. Each sample was rated by three annotators on a scale from 1 to 5, across five distinct metrics. The specifics of the survey design are elaborated in Appendix~\ref{sct: human annotation}.

        The results, presented in \autoref{tab: dataset_human_quality}, confirm the high quality of both datasets. Notably, the scores for ``Coherence'' and ``Fluency'' are impressive, a reflection of the human-crafted nature of the source summaries. \DCNNDM{} outperforms \DWIKIHOW{} in terms of ``Consistency'' and ``Relevance'', which could be attributed to inherent limitations in the original WikiHow dataset as it is observed that \DWIKIHOW{} summaries sometimes either lack necessary details or include extraneous information.

        Regarding ``Aspect Quality'', \DCNNDM{} showcases more clearly defined and distinct aspects compared to \DWIKIHOW{}. The aspects in \DCNNDM{} were noted by the annotators for their clarity, whereas \DWIKIHOW{} was marked by a certain vagueness in aspect demarcation, leading to overlaps in the content of summaries. This variance is likely due to the differing methodologies used in aspect generation for each dataset. Annotators also noted that their ratings for ``Aspect Quality'' were relatively more objective than other criteria. However, they expressed some reservations about the certainty of these scores, indicating the inherent subjectivity in such evaluations.

        \begin{table}[!hbt]
    \setlength{\tabcolsep}{0.5pt}
    \footnotesize
    \begin{tabular}{lccccc}
        \toprule
                  & Coherence & Consistency & Fluency & Relevance & Aspect\\
        \midrule
        CnnDM   & 4.8 (0.4)&4.8 (0.4)   &5.0 (0.2)  & 4.9 (0.3)  & 4.8 (0.8)    \\
        WikiHow & 4.3 (0.9)&3.7 (1.2)   &4.2  (1.0)  & 3.8 (1.3)   &3.9 (1.3)       \\
        \bottomrule
    \end{tabular}
    \caption{Average (std) human evaluation ratings (1--5 scale) on the five quality criteria, determined by 30 instances from the test samples. ``Aspect'': Aspect-Quality, ``CnnDM'':\DCNNDM{} and ``WikiHow'':\DWIKIHOW{}}\label{tab: dataset_human_quality}
\end{table}

        We also assessed the effect of sentence shuffling in the datasets. According to annotator feedback, the disordered texts in \DCNNDM{} did not substantially hinder the identification of topics or aspects. In contrast, \DWIKIHOW{} presented greater challenges in comprehending aspects when sentences were shuffled, exacerbated by the pre-existing ambiguity of the aspects. Thus, for human annotators at least, disordered textual environments increase the complexity of the task, particularly when aspects or topics are not clearly defined.

\section{Baseline Models}\label{sct: Baseline Models}
    In this section, we outline the baseline approaches developed to address the unique challenges presented by our \DABS{} dataset, particularly focusing on aggregating aspect-based information from disordered texts. We explore two primary strategies: (1) Clustering-Based-Summarization (\Cluster{}), which involves clustering sentences by aspects before summarization; and (2) Keyword-Guided (\Keyword{}), utilizing aspect-related keywords to guide the summarization process. Additionally, the Prompt-Guided Summarization (\Prompting{}) is examined, leveraging large-language models for aggregating aspect-based information and generating summaries.

    Both the \Cluster{} and \Keyword{} methods employ BERTopic~\cite{egger2022topic} for sentence clustering and keyword generation, taking advantage of its capabilities in unsupervised clustering and topic modeling. The hyperparameters of BERTopic models are fine-tuned to align the number of references and generated summaries in validation datasets, with further details provided in Appendix~\ref{sct: BERTopic_tuning}. Concurrently, the \Prompting{} method utilizes prompts to direct large-language models, such as GPT-3.5, in automatically aggregating aspect-based information and crafting summaries. Based on previous study~\cite{guo2023length}, we control the length of the generated summaries of all models for mitigating the influence of the generated summary length.

    \subsection{Clustering-Based-Summarization}
        The \Cluster{} method, inspired by previous work~\cite{hayashi2021wikiasp,amar2023openasp}, employs BERTopic to cluster sentences based on their respective aspects. Unlike their supervised approach, we utilize an unsupervised model within BERTopic to handle dynamic aspects with varying numbers and contents. Post-clustering, an abstractive summarization model based on Longformer-Encode-Decoder~\cite{beltagy2020longformer} is used for generating summaries. This summarization model is fine-tuned using original training sets formatted as one-document-to-one-summary, which may potentially inflate its performance.

    \subsection{Keyword-Guided Summarization}

        The \Keyword{} approach, drawing on insights from previous studies in ABS and DABS \cite{ahuja2022aspectnews,he-etal-2022-ctrlsum,yang-etal-2023-oasum}, directs a conventional summarization model to generate aspect-based summaries. This is achieved by incorporating keywords related to each aspect from the source article. For the extraction of unsupervised keywords, we employ C-TF-IDF from BERTopic \cite{grootendorst2022bertopic}, identifying the ten most pertinent keywords for each aspect. These keywords, denoted as \textit{K}, are then prefixed to the shuffled source document \textit{D}, with a ``[SEP]'' token separating them (for example, \textit{K} [SEP] \textit{D}). It's important to note that the number of aspects must be specified in BERTopic during the training phase to facilitate keyword generation. For the \Keyword{} method, we leverage the Longformer-Encode-Decoder model \cite{beltagy2020longformer} as it demonstrates superior performance in previous DABS benchmarks and can accommodate inputs up to 16k characters.

    \subsection{Prompt-Guided Summarization}

    The \Prompting{} method employs GPT-3.5\footnote{Our choice of GPT-3.5 over GPT-4 was influenced by budget constraints and the latter's context length limit of 4k tokens}, a leading large language model, for the \DABS{} task. Given its proven efficacy in ABS tasks with pre-defined aspects~\cite{yang2023exploring}, we propose that well-designed prompts can effectively direct GPT-3.5 in collating aspect-based information and producing summaries. The specifics of these prompts are outlined in Appendix~\ref{sct: GPT-prompts}. To circumvent the context length limitations of GPT-3.5, we utilize different variants of the model suited to each dataset: GPT-3.5-turbo-16k for \DCNNDM{} and GPT-3.5-turbo for \DWIKIHOW{}, selected for their best performance within the respective context length boundaries.

\section{Evaluation and Results}.

    We assessed the performance of our baseline models, as detailed in Section~\ref{sct: Baseline Models}, employing both automatic (Section~\ref{sct: automatic evaluation metrics}) and human evaluation methods (Section~\ref{sct: human evaluation metrics}). Due to the context length constraints of these baseline models, we truncated both the source articles and their corresponding aspect-specific summaries according to their average lengths and standard deviations. Summaries that did not incorporate any sentences from their source articles were omitted from the reference set.

    While fine-tuning our model, we imposed a maximum limit on the number of aspects. However, this limit was not enforced during the evaluation phase. Consequently, baseline models could face penalties for inaccuracies in predicting the correct number of aspects. We have provided additional information about the data preprocessing specific to each dataset in Appendix \ref{sct: Data_process}.

    \subsection{Automatic Evaluation Results}\label{sct: results}
        The automatic evaluation encompassed both summary quality and aspect identification accuracy. For \Keyword{} and \Cluster{}, multiple experiments were conducted using varied random seeds. For \Prompting{}, a small-scale study identified the most effective prompt, with details in Appendix~\ref{sct: hyper-and-seed}.

        \begin{table*}[!htb]
        \centering
        \begin{tabular}{lllllll}
        \toprule
        Model    & Dataset                    & \AspDiff{}    & BERTScore           & Rouge-1             & Rouge-2            & Rouge-L             \\ \midrule
        \Keyword{} & \multirow{3}{*}{\DCNNDM{}}     & 1.3 (0.0)         & 14.2 (0.1)         & 24.8 (0.0)         & 8.9 (0.2)          & 17.2 (0.1)         \\
        \Cluster{}  &                            & 1.3 (0.0)         & 15.1 (0.2)         & 25.4 (0.1)         & 9.1 (0.1)          & 17.1 (0.1)         \\
        \Prompting{}  &                            & 6.2              & 9.1               & 12.4               & 4.1               & 9.1                    \\
        \midrule
        \Keyword{} & \multirow{3}{*}{\DWIKIHOW{}}   & 2.7 (0.1)         & 30.5 (0.1)         & 14.6 (0.2)         & 5.0 (0.1)         & 14.2 (0.2)         \\
        \Cluster{}  &                            & 2.7 (0.1)         & 31.3 (0.0)         & 19.1 (0.1)         & 7.8 (0.1)         & 18.5 (0.1)         \\
        \Prompting{}  &                            & 5.5                & 17.7                & 11.4                & 3.8                & 10.3                \\
        \bottomrule
        \end{tabular}
        \caption{The performance of baselines across \DCNNDM{} and \DWIKIHOW{}. Mean scores are reported, accompanied by standard deviations in brackets. Due to budgetary constraints, the results for GPT-3.5 are derived from a single experimental run.}\label{tab: automatic_main_results}
    \end{table*}

        The results, shown in \autoref{tab: automatic_main_results}, indicate that the two-step methods (\Keyword{} and \Cluster{}) generally outperform \Prompting{} across all metrics, particularly in \AspDiff{}. This trend suggests that \Prompting{} might struggle to accurately discern the nuances of different aspects within samples. Notably, \Cluster{} exhibits superior performance over \Keyword{}, suggesting that simply providing automatic generation of aspect-related keywords is insufficient for effectively extracting aspect-based information from disordered texts.

        A pilot check was conducted to understand the disparity in aspect identification between \Prompting{} and the other baselines. This check reveals that \Prompting{} frequently splits a single topic or aspect into multiple segments, especially in \DCNNDM{} samples with one or two topics. This indicates that \Prompting{}'s zero-shot in-context learning approach may have difficulties in recognizing the cohesion of a single topic or aspect. Conversely, \Cluster{} and \Keyword{}, leveraging BERTopic for aspect identification, appear to have a more accurate grasp of topic or aspect content.

    \subsection{Human Evaluations}

    In our study, human evaluations complemented automatic metrics by assessing the performance of models on the \DCNNDM{} and \DWIKIHOW{} subsets. We selected thirty instances from each dataset for evaluation by three judges, based on the criteria outlined in Section~\ref{sct: human evaluation metrics} and using the interfaces described in \autoref{sct: human annotation}, with results detailed in \autoref{tab: human_main_results}.
    
    \begin{table*}
    \centering
    \small
    \begin{tabular}{llllllll}
    \toprule
        Dataset                    & Model    & Coherence   & Consistency & Fluency     & Relevance   & Aspect-Quality & Rank        \\
        \midrule
        \multirow{3}{*}{\DCNNDM{}}   & \Keyword{} & 3.17 (0.76) & 2.79 (0.93) & 3.38 (0.97) & 3.25 (0.68) & 3.00 (1.14)    & 2.21 (0.83) \\
                                     & \Cluster{}  & 3.50 (0.72) & 3.21 (0.78) & 3.50 (0.72) & 3.46 (0.88) & 4.04 (1.00)    & 1.71 (0.69) \\
                                     & \Prompting{}  & 3.21 (1.28) & 2.71 (0.75) & 4.29 (1.12) & 3.12 (0.80) & 4.25 (1.07)    & 2.08 (0.88) \\
        \midrule
        \multirow{3}{*}{\DWIKIHOW{}} & \Keyword{} & 2.16 (1.29) & 2.58 (0.72) & 2.48 (1.03) & 1.90 (0.94) & 1.97 (1.35)    & 2.68 (0.48) \\
                                     & \Cluster{}  & 2.74 (1.18) & 3.23 (0.99) & 3.06 (1.03) & 2.84 (1.00) & 2.55 (1.29)    & 2.32 (0.73) \\
                                     & \Prompting{}  & 4.55 (0.57) & 3.58 (1.09) & 4.39 (0.72) & 4.10 (1.14) & 4.58 (0.81)    & 1.00 (0.00) \\
        \bottomrule
        \end{tabular}
        \caption{Average (std) human evaluation ratings (1--5 scales) on the five quality criteria and the ranking (1--4 scales), determined by 30 instances from the test samples.}\label{tab: human_main_results}
\end{table*}

    The evaluations revealed diverse performances across datasets and evaluation criteria. In the \DCNNDM{} dataset, \Prompting{} excelled in two specific criteria, whereas \Cluster{} was superior in three. For the \DWIKIHOW{} dataset, \Prompting{} outperformed all other baselines across all criteria, indicating its greater effectiveness in less aspect-distinct scenarios. Consistent with automatic evaluations, \Cluster{} generally surpassed \Keyword{} in performance.
    
    Notably, \Prompting{} was consistently rated highest for ``Fluency'' and Aspect Quality.'' This is in contrast to automatic evaluations, where \Prompting{} was penalized for generating fewer aspects. Further inquiry with judges revealed that \Prompting{} tends to fragment one aspect into several, accounting for this discrepancy. Additionally, annotators' preference for a higher number of aspects in \DWIKIHOW{}, as opposed to reference standards, highlights the subjective nature of determining the optimal number of aspects, especially in nuanced aspect differentiation.

\section{Ablation Analysis}\label{sct: GPT-3.5-experiments} 
  Our ablation study delves into three key areas: 1) Exploring few-shot learning capabilities of GPT-3.5; 2) Assessing the impact of disordered text settings on model performance; and 3) Examining the effects of variations and discrepancies in aspect number.
    
        \begin{table*} [!bt]
        \centering
        \begin{tabular}{lllllll}
        \toprule
        Model                   & \#Sample & \AspDiff{} & BERTScore & Rouge-1 & Rouge-2 & Rouge-L \\ \midrule
        zero-shot    & 0    & 5.6                       & 18.3                   & 11.7                    & 4.0                     & 10.6                    \\ \midrule
        \multirow{3}{*}{In-context}& 1    & 4.1                       & 26.6                   & 16.6                    & 5.9                     & 15.5                    \\
                                & 3    & 3.7                       & 29.0                   & 18.1                    & 6.7                     & 17.0                    \\
                                & 6    & 3.5                       & 29.6                   & 18.6                    & 7.0                     & 17.5                    \\ \midrule
        \multirow{3}{*}{Fine-tuning} & 50   & 4.2                       & 17.1                   & 11.0                    & 4.3                     & 10.2                    \\
                                & 100  & 3.4                       & 25.4                   & 16.1                    & 6.7                     & 15.3                    \\
                                & 200  & 4.6                       & 15.1                   & 10.1                    & 4.1                     & 9.4                     \\ \bottomrule
        \end{tabular}
        \caption{Comparison of zero-shot learning, in-context learning, and fine-tuned \Prompting{} based on sampled data.  ``\#Samples'': the number of samples used.}\label{tab: gpt-3.5-experiment}
    \end{table*}

    \subsection{Few-shot Settings for GPT-3.5}\label{sct: GPT-3.5-tuning}
      The results for \Prompting{}, as detailed in \autoref{tab: automatic_main_results}, were obtained using zero-shot learning. To further assess the capabilities of few-shot in-context learning and fine-tuning, we conducted additional experiments with the \DWIKIHOW{} dataset. These experiments were not extended to the \DCNNDM{} dataset due to the constraints of GPT-3.5-turbo-16k, specifically its limitations on fine-tuning and the context length restrictions applicable to few-shot in-context learning.

        Table \ref{tab: gpt-3.5-experiment} shows that few-shot learning and fine-tuning can enhance the performance of GPT-3.5 by minimizing the absolute differences in aspect counts between the predicted and ground truth summaries \footnote{Samples are shown in Appendix \ref{sct:examples-GPT-3.5}}. This implies that GPT-3.5 still requires domain-specific information to accurately differentiate between various aspects or topics. Interestingly, we observe that few-shot learning outperforms fine-tuning when the sample size is limited, suggesting that it may be a more cost-effective approach under similar conditions.

    \subsection{The Impact of Disorder Texts}

        Here, we investigate how disordered text environments, a central aspect of the \DABS{} dataset, affect summarization model performance. By conducting control experiments with organized texts under identical experimental setups as previously outlined in Section~\ref{sct: results}, the study reveals varied impacts on model efficacy across different methods and datasets (see \autoref{tab: automatic_organized_results}). 
        
        \begin{table*}[!htb]
        \centering
        \begin{tabular}{lllllll}
        \toprule
        Model    & Dataset                    & \AspDiff{}    & BERTScore           & Rouge-1             & Rouge-2            & Rouge-L             \\ \midrule
        \Keyword{} & \multirow{3}{*}{\DCNNDM{}}     & 1.2 (0.0)         & 2.0 (-12.2)         & 18.7 (-6.1)         & 5.3 (-3.6)          & 12.7 (-4.5)         \\
        \Cluster{}  &                            & 1.2 (0.0)         & 19.0 (2.8)         & 29.0 (2.6)         & 12.4 (2.3)          & 20.5 (2.4)         \\
        \Prompting{}  &                            &5.6 (-0.6)              & 12.3 (3.2)              & 16.9 (4.6)              & 6.0 (1.9)               & 11.9 (2.8)                    \\
        \midrule
        \Keyword{} & \multirow{3}{*}{\DWIKIHOW{}}   & 2.7 (0.0)         & 27.5 (-3.0)         & 15.9 (1.3)         & 6.0 (1.0)         & 15.3 (1.1)         \\
        \Cluster{}  &                            & 2.7 (0.0)         & 29.9 (-1.4)         & 18.2 (-0.9)         & 7.6 (-0.2)         & 17.6 (-0.9)         \\
        \Prompting{}  &                            & 4.8 (-0.7)                & 20.5 (2.8)                & 13.3 (1.9)                & 4.7 (0.9)                & 12.1 (1.8)                \\
        \bottomrule
        \end{tabular}
        \caption{The performance of baseline models on the \DCNNDM{} and \DWIKIHOW{} sub-datasets, noting the effects of organized settings in parentheses. For the ``\AspDiff{}'' metric, a negative value signifies an improvement, whereas the opposite is true for the remaining four metrics.}\label{tab: automatic_organized_results} 
    \end{table*}

        Specifically, the \Prompting{} approach showed uniform improvement in performance metrics for both the \DCNNDM{} and \DWIKIHOW{} datasets when applied to organized texts. In contrast, the \Keyword{} and \Cluster{} approaches demonstrated mixed results, with the former seeing a decline in performance on the \DCNNDM{} dataset but an improvement on \DWIKIHOW{} (with an exception for the \AspDiff{} metric), while the latter improved on \DCNNDM{} but deteriorated on \DWIKIHOW{}. These outcomes underscore the complex nature of disordered texts' impact on summarization tasks, suggesting the need for deeper exploration into how text structure affects summarization model performance.

    \subsection{The Influence of Aspect Number Variations and Discrepancies}
        Given the significance of aspect count in \DABS{}, we investigate how varying aspect numbers impact the performance of the models.

        \begin{figure*}[!thb]
            \centering
            \includegraphics[width=1.95\columnwidth]{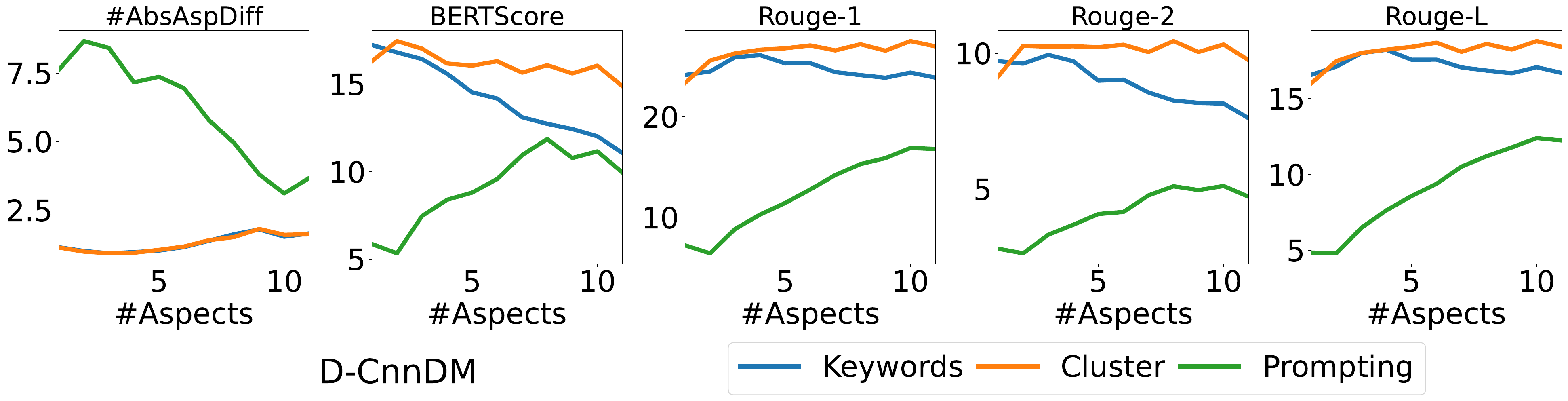}
            \includegraphics[width=1.95\columnwidth]{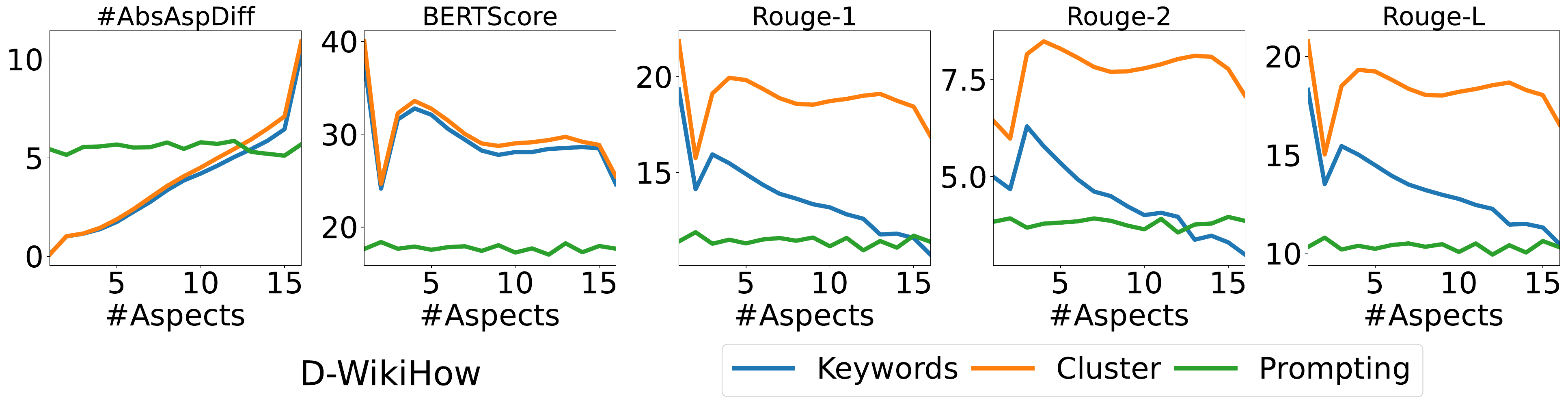}
            \caption{Performance variation of baselines with changes in aspect number in the reference. The final data point represents cases where the number of aspects in the reference summary equals or exceeds 12 for \DCNNDM{} and 16 for \DWIKIHOW{}.}\label{fig: aspect-influence}
                 \vspace{-5mm}
        \end{figure*}

        \autoref{fig: aspect-influence} demonstrates how the performance of various baseline models is affected by changes in the number of aspects across two datasets. In the \DCNNDM{} dataset, \Prompting{} improves as the number of aspects increases, whereas \Keyword{} and \Cluster{} show a decline in performance. Conversely, in the \DWIKIHOW{} dataset, the number of aspects has a negligible effect on \Prompting{} but adversely affects \Keyword{}. These observations highlight that the impact of aspect numbers varies significantly across different models and datasets.

        \begin{table}[!htb]
        \centering
        \setlength{\tabcolsep}{1.1pt}
        \small
        \begin{tabular}{llllll}
        \toprule
        Model                     & Dataset         & BS       & R1       & R2       & RL       \\ \midrule
        \Keyword{}   & \multirow{3}{*}{\DCNNDM{}}   & 19.4(5)  & 32.1(7)  & 11.9(3)  & 22.4(5)  \\
        \Cluster{}   &                              & 20.5(5)  & 33.0(8)  & 12.4(3)  & 22.4(5)  \\
        \Prompting{} &                              & 21.1(12) & 26.4(14) & 9.7(6)   & 19.5(10) \\ \midrule
        \Keyword{}   & \multirow{3}{*}{\DWIKIHOW{}} & 51.5(21) & 29.8(15) & 11.9(7)  & 29.0(15) \\
        \Cluster{}   &                              & 55.0(24) & 38.1(19) & 18.2(10) & 37.0(19) \\
        \Prompting{} &                              & 35.5(18) & 26.4(5)  & 12.5(9)  & 24.3(14) \\ \bottomrule
        \end{tabular}
        \caption{The performance of all baselines across datasets with ``perfect-match''. Mean scores are reported, accompanied by the improvement caused by ``perfect-match'' settings in brackets. ``BS'':``BERTScore'',``R1'':``Rouge-1'',``R2'':``Rouge-2'',``RL'':``Rouge-L''.}\label{tab: automatic_no-padding_results}
             \vspace{-5mm}
    \end{table}

        To further assess if our model's improved performance is due to its ability to accurately predict the number of aspects or the quality of the generated summaries, we experimented by matching the reference and generated summary numbers (``perfect-match''). For each sample, we consider only the top-k aspects where each model achieves optimal performance, and $k$ is the smallest aspect number among all models and the reference. The results, displayed in \autoref{tab: automatic_no-padding_results}, show significant improvement for all baselines, particularly for \DWIKIHOW{}. This suggests that identifying and combining aspects in \DWIKIHOW{} presents more challenges than in \DCNNDM{}, likely due to the methodologies employed for aspect generation and the inherent ambiguity in defining aspects.

\section{Conclusion}

    Our \DABS{} benchmark introduces the first dataset dedicated to dynamic aspect-based summarization in disordered texts, innovatively crafted from high-quality summaries of existing datasets. This approach sidesteps the conventional challenge of manual summary generation by transforming available datasets for summarization.

    The evaluation of leading-edge baseline models has exposed a notable deficiency in the current capacity to effectively tackle this complex task. Our analysis shows that contemporary models struggle significantly with the nuances of dynamic aspect-based summarization in disorganized texts, underscoring the task's complexity and highlighting avenues for future research and enhancement.

\section{Limitations}
While not raising significant ethical concerns, our study acknowledges limitations associated with the use of publicly available datasets, which may contain sensitive or potentially offensive content. We recommend caution in handling these datasets. The transformation of existing summaries to create dynamic aspect-based summaries in disordered texts, despite manual checks, may not ensure uniform quality across all samples, prompting users to verify anomalies in further experiments.

The \DABS{} benchmark, derived from two distinct datasets, offers variety in domains and granularity but does not cover the entire spectrum of dynamic aspect-based summarization challenges in disordered texts. It serves as a baseline for general domains, with a recommendation to adapt our methodology for domain-specific benchmarks in dynamic aspect-based summarization.

Furthermore, our reliance on GPT-3.5-turbo introduces potential biases due to undisclosed training data, possibly including our test data, which could influence our findings. The prompt-dependent nature of our experiments with GPT-3.5-turbo (the specific prompts utilized are detailed in the Appendix) underscores that our conclusions about its effectiveness are contingent on the specific prompts used, indicating that different prompts could yield varied results. This highlights the need for transparent model documentation and the exploration of prompt-independent evaluation methods to enhance the reliability of findings in dynamic aspect-based summarization research.

\bibliography{custom}

\setcounter{table}{0}
\setcounter{figure}{0}
\renewcommand\thefigure{\Alph{section}\arabic{figure}}
\renewcommand\thetable{\Alph{section}\arabic{table}}
\clearpage
\appendix

\section{Dataset}\label{sct: Dataset-appendix}
    \begin{table*}[!htb]
    \centering
    \setlength\tabcolsep{4pt}
    \begin{tabular}{llllllll}
    \toprule
    Dataset   & Domain    & \#Train   & \#Valid & \#Test & Avg.Arc. Len & Avg.Sum. Len & Avg.\# Asp  \\ \midrule
    AnyAspect & News      & 287,227   & 13,368  &11,490  & 681 (332)    & 20 (18)      & 7.06 (4.20) \\
    OASUM     & Wikipedia & 1,937,776 &60,526   &60,481  & 792 (1,181)  & 35( 37)      & 1.8 (1.48)  \\
    OpenASP   & News      & 476       & 238     & 596    & 7,930        & 96           & 3.1         \\
    ENTSUMV2  & News      & N/A       & N/A     & N/A    & 1,002        & 46           & N/A         \\ \bottomrule     
    \end{tabular}
    \caption{Statistics of the datasets. ``Avg.Arc.Len'', ``Avg.Sum.Len'', and ``Avg.\#Asp'' denote the average number of words of articles, summaries, and the average number of aspects per sample. Standard deviations are provided in brackets. Since the dataset ENTSUMV2 is not publicly available, we only report the statistics in their paper.
    }\label{tab: dabs_statistics}
\end{table*}
    \subsection{\DABS{} Distribution Statistics}
\begin{figure}[!thb]
            \centering
            \includegraphics[width=0.95\columnwidth]{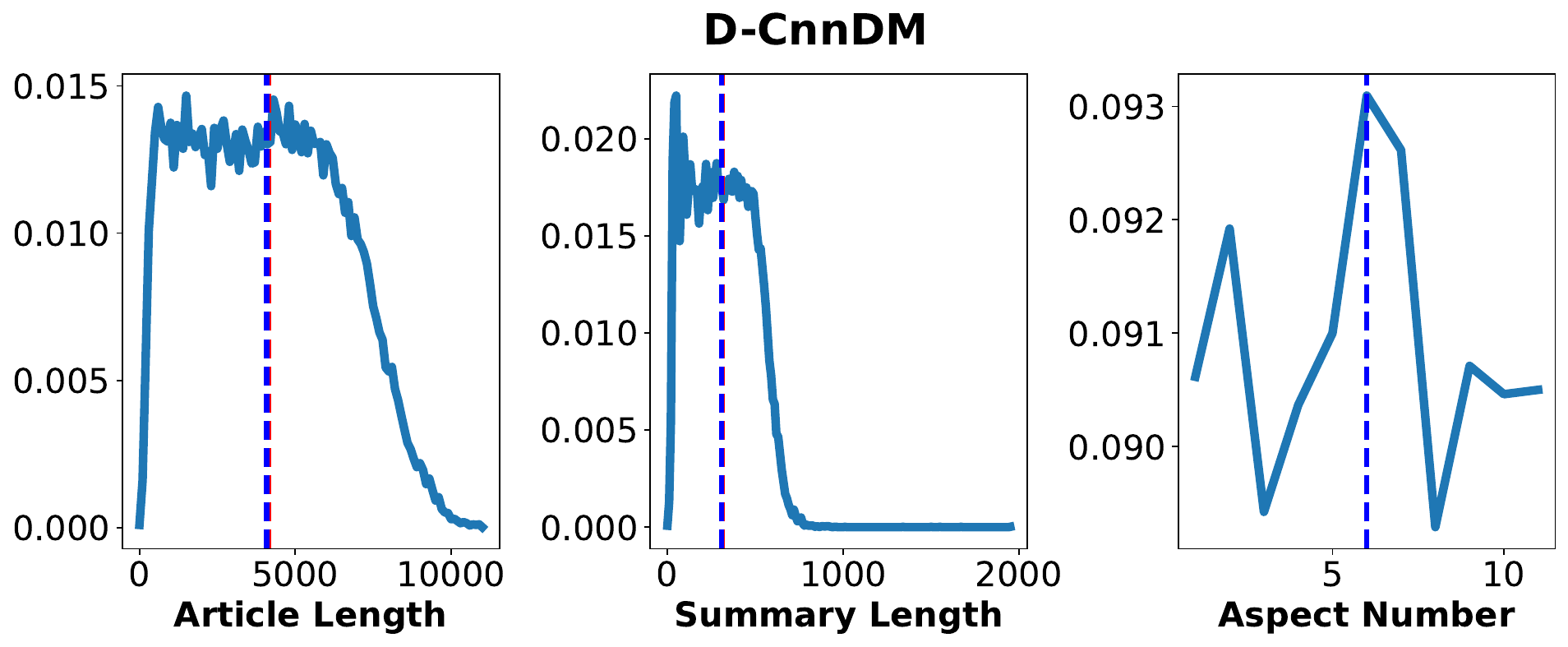}
            \includegraphics[width=0.95\columnwidth]{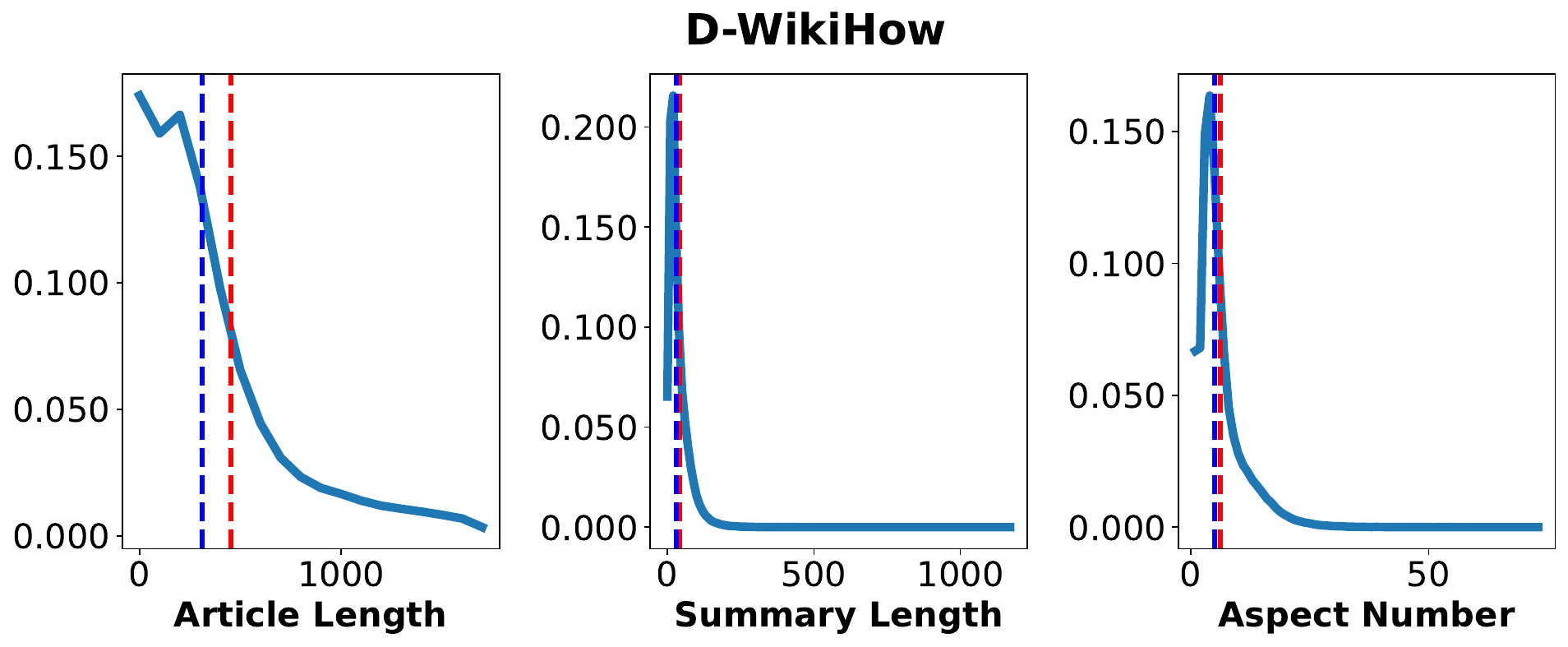}
            \caption{Distributions of article length, summary length, and aspect number across datasets. Mean and median values are marked with red and blue lines, respectively. For clarity, distributions are shown within three standard deviations from the mean for source articles and references.}\label{fig: data-distribution}
        \end{figure}

        In this appendix, we analyze the distribution of various statistics in our datasets, as shown in \autoref{fig: data-distribution}. For \DCNNDM{}, the article lengths approximate a normal distribution, while the summary lengths follow a long-tail distribution. In contrast, for \DWIKIHOW{}, both article and summary lengths exhibit long-tail distributions.
        
        The aspect number distribution in \DWIKIHOW{} shows a right-skewed pattern, whereas, in \DCNNDM{}, it is almost uniform. This difference allows us to investigate how aspect number distribution influences dynamic aspect-based summarization. Given these variations, \DCNNDM{} may offer a wider variety of aspects due to its balanced distribution.
        
    \subsection{\DABS{} Samples}
        \label{sct: DABS-samples}
\begin{figure*}[!htb]
            \centering
            \includegraphics[width=1.95\columnwidth]{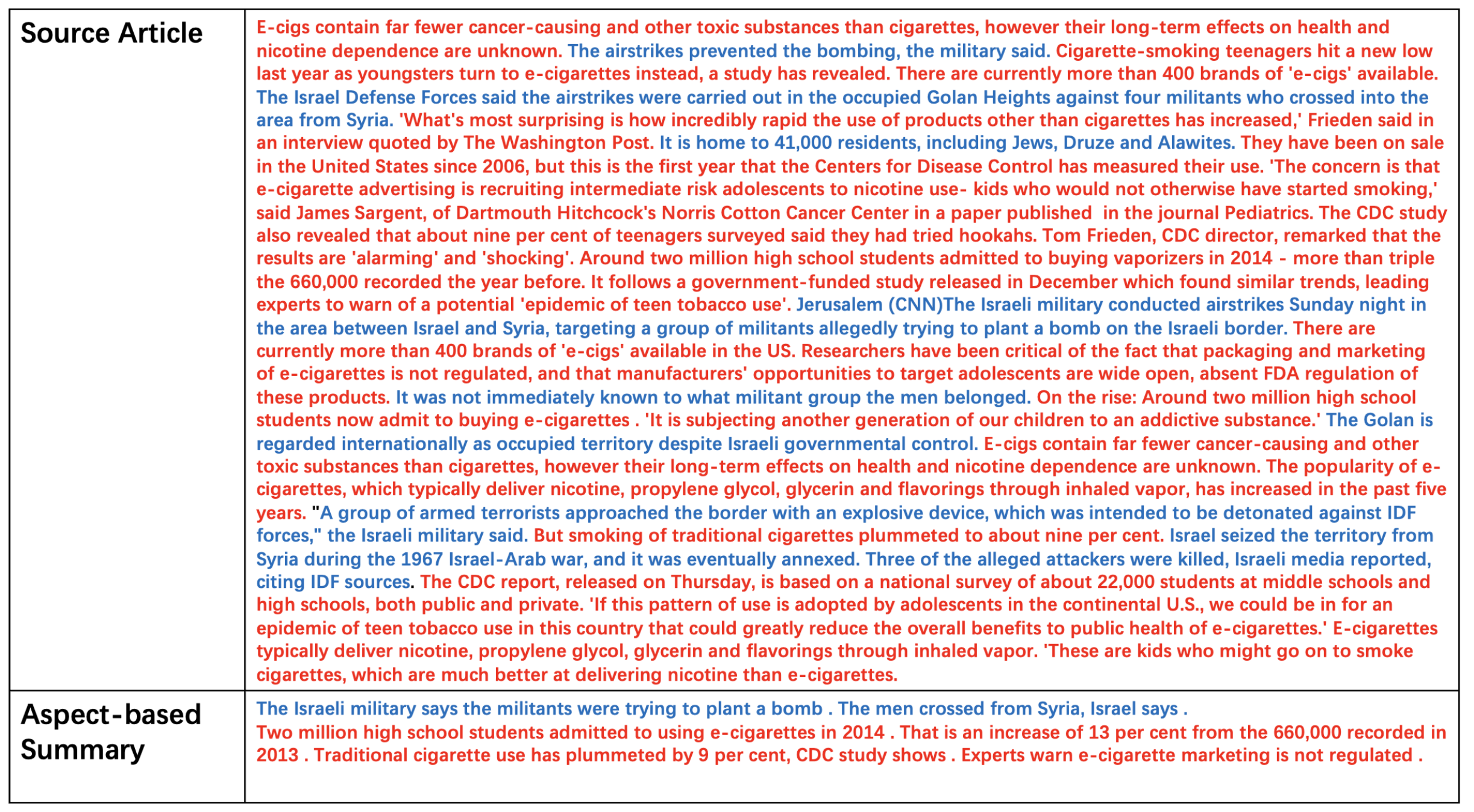}
            \includegraphics[width=1.95\columnwidth]{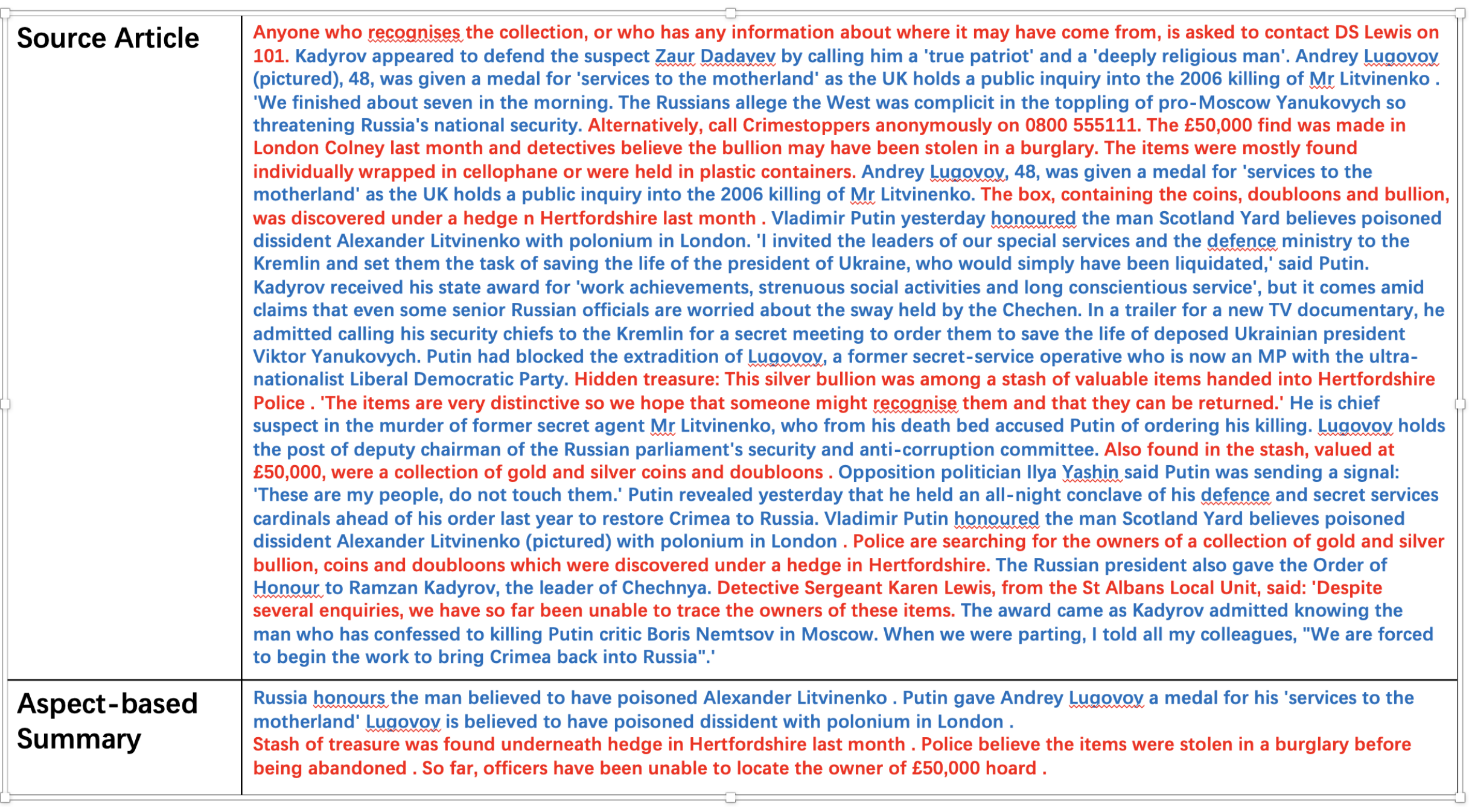}
            \caption{Examples from the \DCNNDM{}{}, illustrating various aspects represented by different colors.}\label{fig: D-CnnDM_more_sample}
        \end{figure*}

\begin{figure*}[!htb]
            \centering
            \includegraphics[width=1.95\columnwidth]{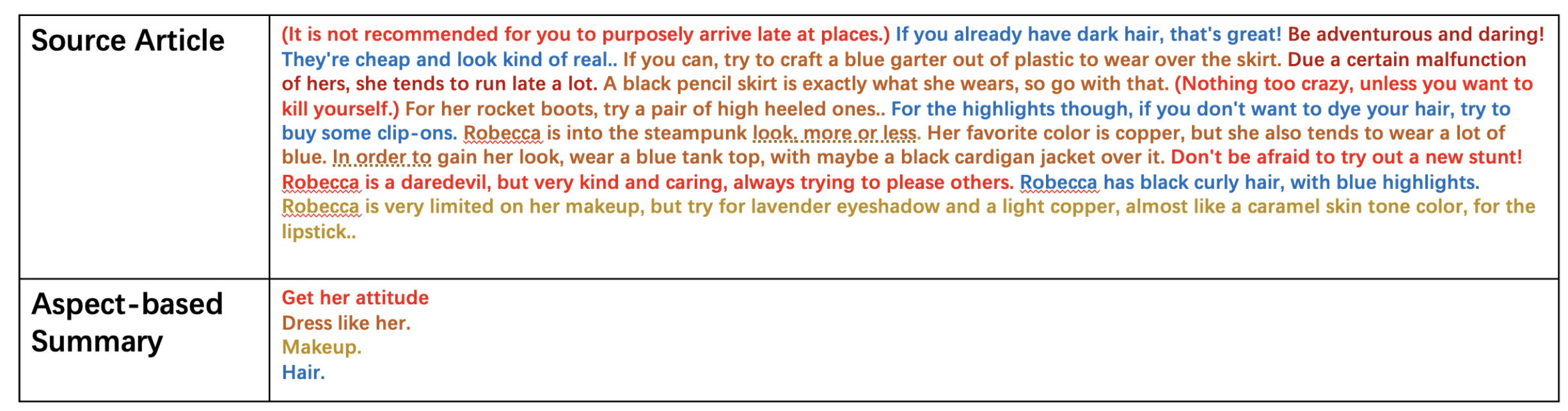}
            \includegraphics[width=1.95\columnwidth]{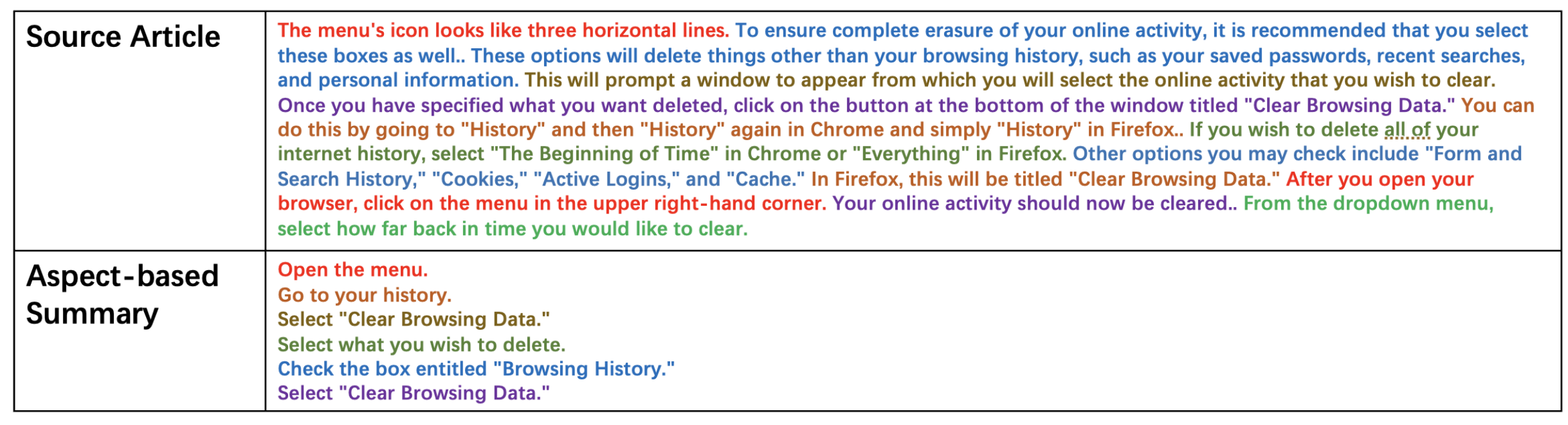}
            \caption{Examples from the \DWIKIHOW{}, illustrating various aspects represented by different colors.}\label{fig: D-Wikihow_more_sample}
        \end{figure*}

         \autoref{fig: D-CnnDM_more_sample}  and \autoref{fig: D-Wikihow_more_sample} show the source articles and aspect-based summaries, which belong to 4 randomly selected different instances from \DABS{} (2 from \DCNNDM{} and 2 from \DWIKIHOW{}).

    \subsection{Details on DABS Datasets}
        \autoref{tab: dabs_statistics} provides detailed information about the datasets previously introduced in \autoref{tab: related_work_comparision}.

\section{Experimental Details}
    \subsection{Computing Infrastructure}\label{sct:computing_infrastructure}
    For our experiments, we used a Lambda machine equipped with 250 GB of memory, 4 RTX 8000 GPUs, and 32 CPU cores. The machine runs on Ubuntu 20.04, and our experiments are conducted using Python 3.8.10. The CUDA version is 11.9, and the GPU Driver Version is 520.61. The main packages we utilize include bertopic (0.14.1), cuml-cu11 (23.4.1), deepspeed (0.8.0), torch (1.13.1), scikit-learn (1.1.2), sentence-transformers (2.2.2), scipy (1.9.1), transformers (4.22.1), and numpy (1.23.3). The complete list of packages will be provided in the code release.

    \subsection{Data Processing}\label{sct: Data_process}
        \autoref{tab: thresholds_info} displays the thresholds for source article length, single-aspect summary length, and the maximum aspect number used for truncation during the experiments. The threshold for the summary length corresponds to the single-aspect length, so the total length of the summaries is multiplied by the aspect number. 

        \begin{table}[!hbt]
    \centering
    \begin{tabular}{llll}
    \toprule
              & Arc. Leng & Sum. Length & \#Asp \\ \midrule
    \DCNNDM{}   & 11,264    & 76          & 12   \\
    \DWIKIHOW{} & 2,040     & 20          & 16   \\ \bottomrule
    \end{tabular}
    \caption{The thresholds for the source article length, single-aspect summary length, and the maximum aspect number for each sample.}\label{tab: thresholds_info}
\end{table}
        
        For \DCNNDM{} and \DWIKIHOW{}, we truncate the source article and single-aspect summaries based on the mean and standard deviation of the length (approximately equal to mean + 2 \(\times \) std).
    
        Regarding the maximum aspect number, we set the thresholds for \DCNNDM{} based on the largest \(\#Asp\) in the datasets (12). For \DWIKIHOW{}, considering it follows a long-tail distribution, we set the threshold to 16 (approximately equal to the mean + 2 \(\times \) std).

    \subsection{Prompts for GPT-3.5}\label{sct: GPT-prompts}
        The prompts employed for GPT experiments adhere to the following template: ``You are a summarizer. Your task is to summarize sentences across multiple aspects or topics. The sentences are shuffled. Please generate multiple summaries, ensuring each summary pertains to a single aspect or topic. Limit each summary to one sentence, exclude the topic name, and restrict its length to fewer than N tokens. Separate each summary with [SEP]''. Here, N represents the output length constraint. For zero-shot inference, the prompt is augmented with the phrase ``There should be no more than M topics or aspects'' serving to restrict the total number of aspects or topics considered. In this context, M denotes the maximum allowable count of topics or aspects.

    \subsection{BERTopic Hyperparameter Tuning}\label{sct: BERTopic_tuning}
        For the baselines, we performed a grid search for tuning the BERTopic model hyperparameters. 
        The hyperparameters tuned for the BERTopic include ``n\_neighbours'', ``n\_component'', and ``min\_dist'' which control the cluster size and the samples within a cluster. We performed BERTopic clustering on each sample of the validation data and selected the combination of hyperparameters that minimized the absolute difference between the generated and reference aspect numbers.

    \subsection{Hyperparameters and Random Seeds}\label{sct: hyper-and-seed}
        For our experiments, we used three random seeds (0, 10, and 42) for the complete dataset experiments.  We used the Hugging Face implementation for fine-tuning the language model with a batch size of 4 due to GPU memory limitations. The training epoch was set to 10 with an early stop, and all other training process hyperparameters were set to the default values provided by the package. We also used DeepSpeed to reduce memory requirements during the fine-tuning process. The specific DeepSpeed configuration will be provided along with our code.

\section{Human Annotation}\label{sct: human annotation}

\begin{figure*}
        \centering
        \includegraphics[width=1.65\columnwidth]{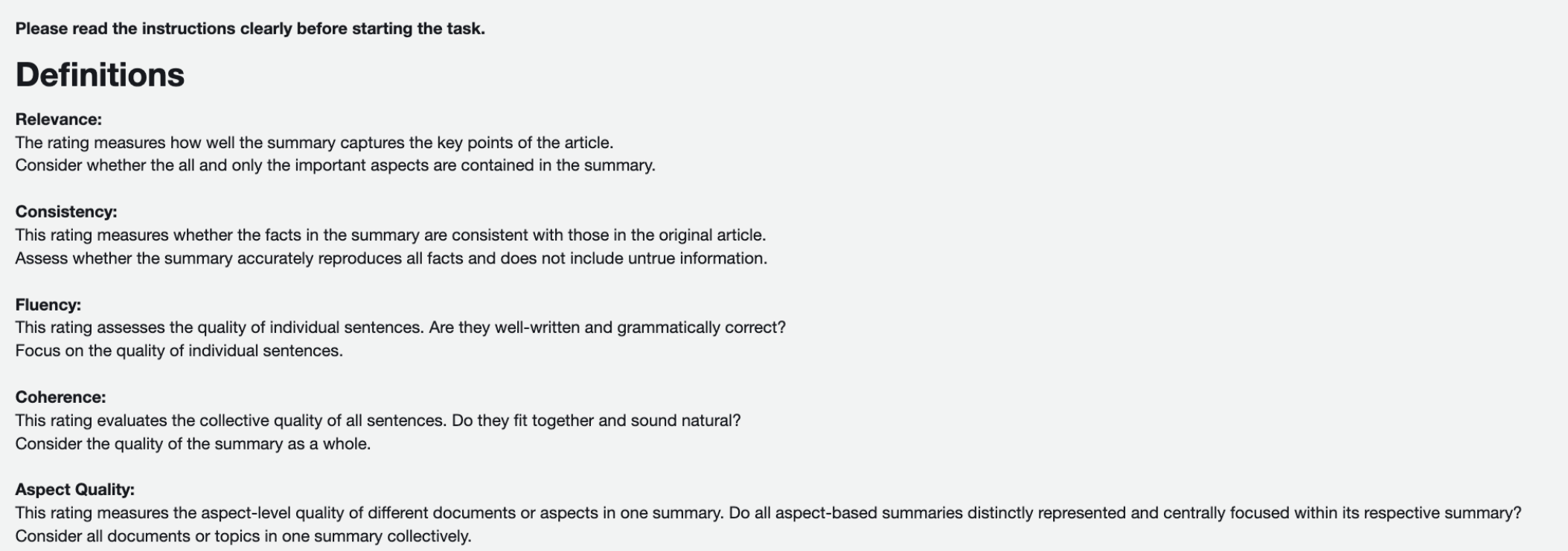}
        \includegraphics[width=1.65\columnwidth]{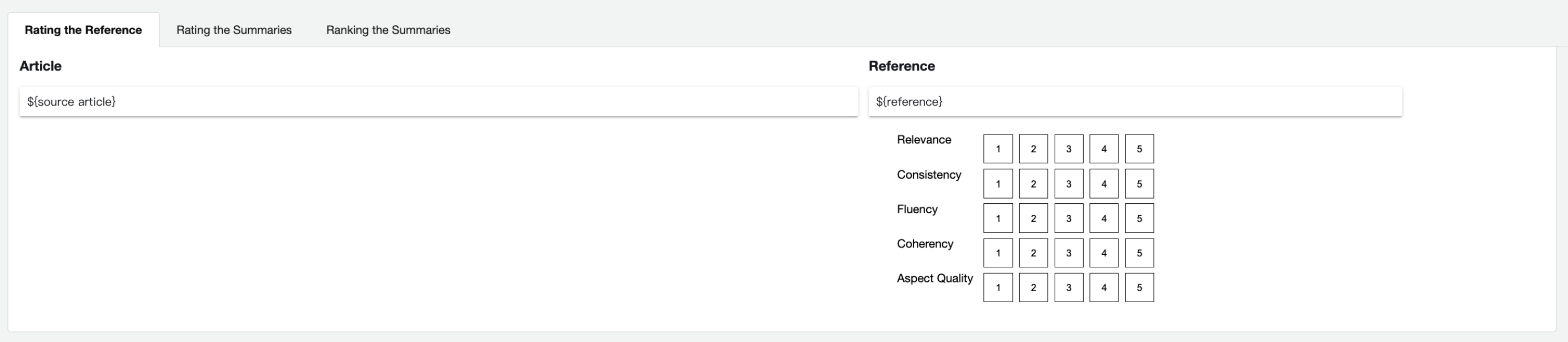}
        \includegraphics[width=1.65\columnwidth]{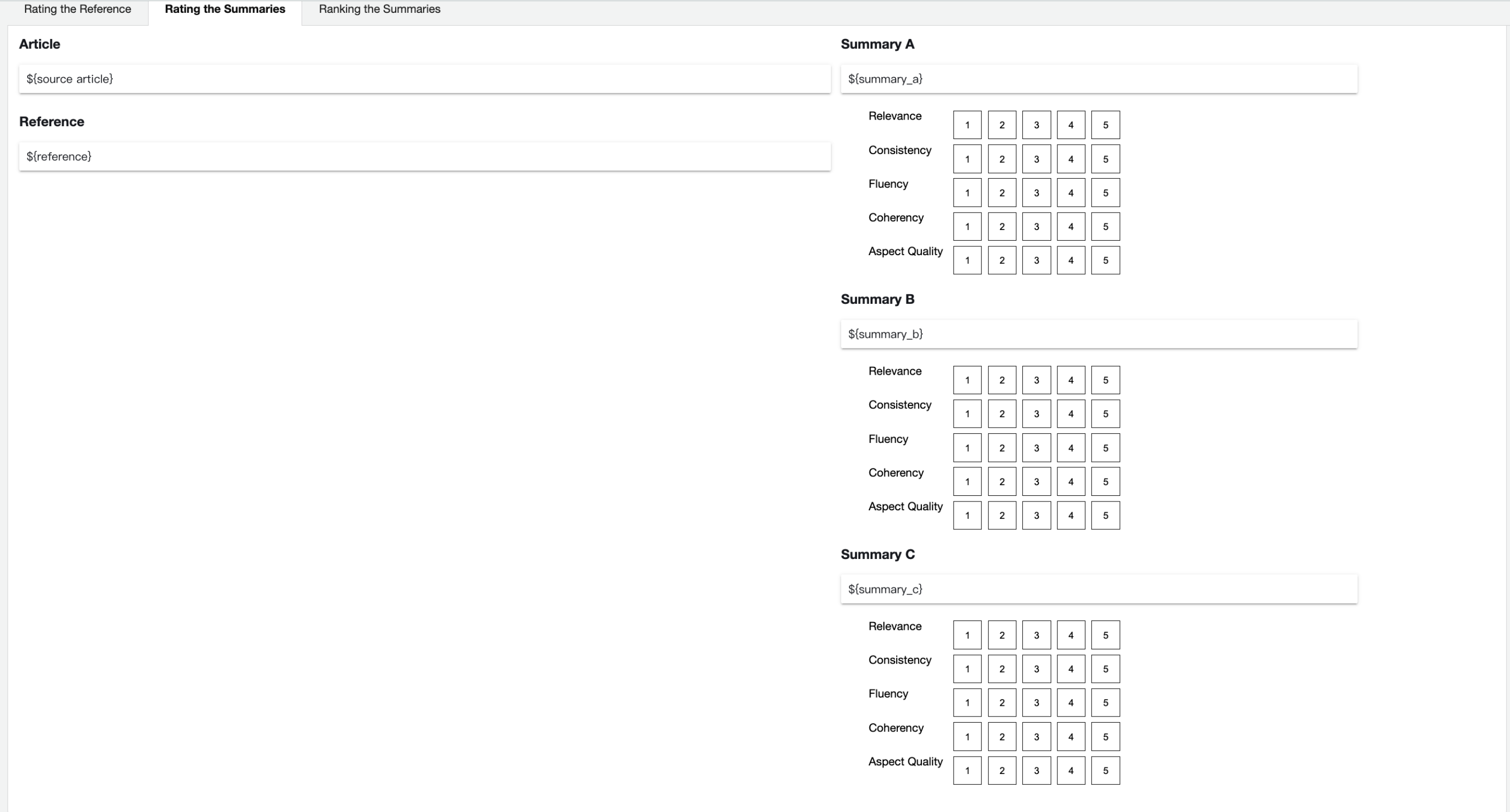}
        \includegraphics[width=1.65\columnwidth]{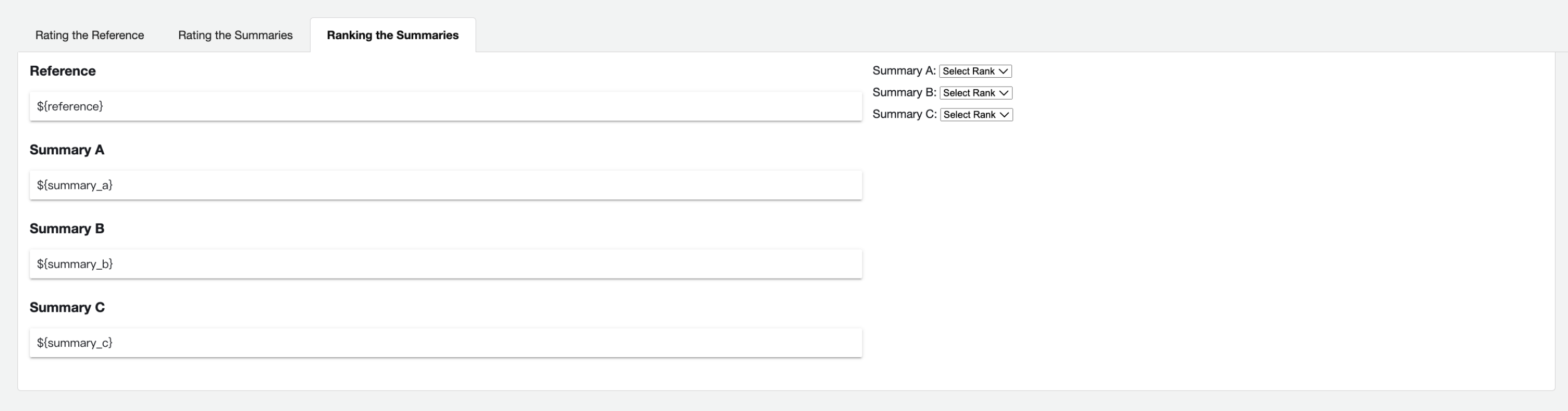}
        \caption{Example of the human annotation interface}\label{fig: human-annotation}
    \end{figure*}

    The human annotation process is detailed in \autoref{fig: human-annotation}, outlining a structured approach for annotators. They are instructed to thoroughly review the source article and its ground truth summary to: (1) Assess the ground truth based on five criteria: \textit{Coherence}, \textit{Consistency}, \textit{Fluency}, \textit{Relevance}, and \textit{Aspect Quality}; (2) Evaluate generated summaries against these criteria, using the ground truth for reference; (3) Rank the summaries according to their overall effectiveness. All annotators possess at least a bachelor's degree and are proficient in English. They are compensated between \$10 to \$15 per hour for their contributions.

\section{Examples of Summaries}
    \subsection{Examples of Main Results}\label{sct: examples-main}
\begin{figure*}[!hbt]
        \centering
        \includegraphics[width=1.95\columnwidth]{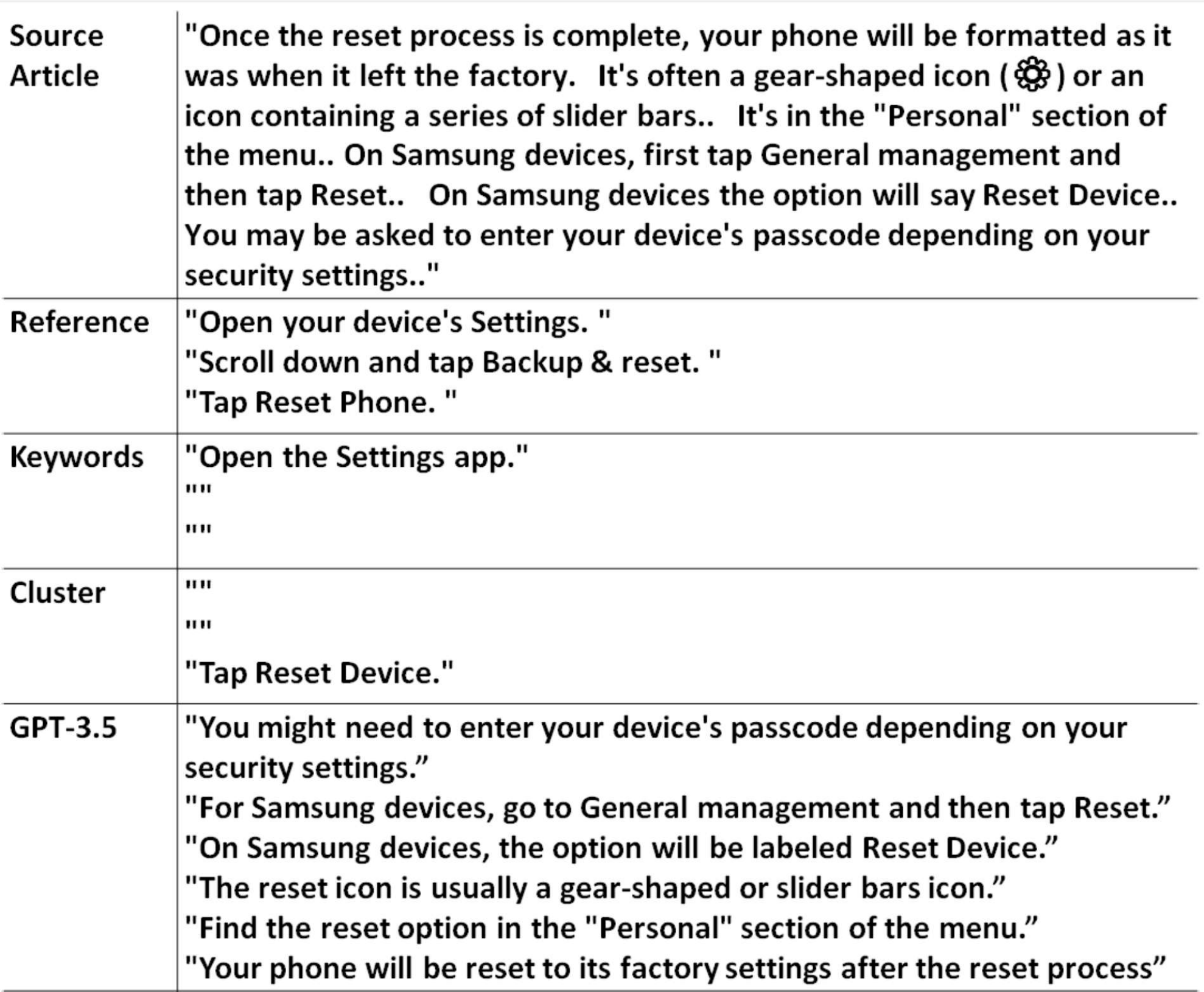}
        \caption{An example of the source article, reference summaries, and generated summaries by baselines. Empty quotes (``\,'') indicate missing generated summaries.}\label{fig: summary_sample}
    \end{figure*}

    \autoref{fig: summary_sample} presents an illustrative example featuring the source article, reference summaries, and summaries generated by all baselines. The source article addresses the topic of sending a friend invitation on Facebook. We have pre-aligned the reference and generated summaries for ease of comparison. The presence of empty quotes (``\,'') signifies the absence of a corresponding generated summary for a given reference summary. Notably, \Prompting{} generates 6 aspect-based summaries while the other two only generate one aspect-based summary with information missing. The difference in aspect discovery preference is in alignment with what we observe in other experiments.
    
    \subsection{Examples of Fine-tuned GPT-3.5}\label{sct:examples-GPT-3.5}
\begin{figure*}[!hbt]
        \centering
        \includegraphics[width=1.95\columnwidth]{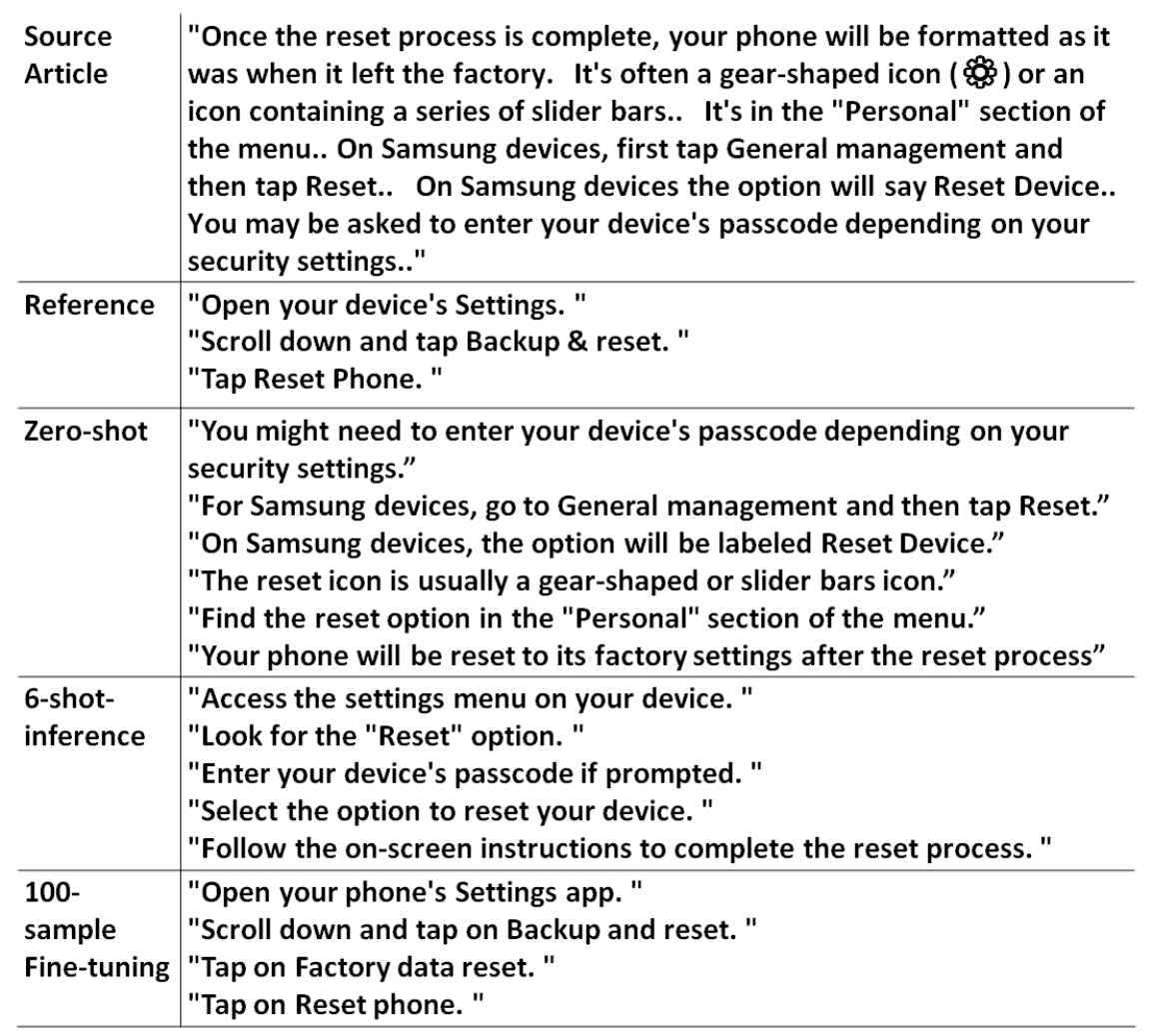}
        \caption{Example of the source article, reference summaries, and \Prompting{} generated summaries with 6-shot inference and 100 samples fine-tuning.}\label{fig: openai_summary_sample}
    \end{figure*}

    \autoref{fig: openai_summary_sample} presents an illustrative example featuring the source article, reference summaries, and summaries generated by \Prompting{} with the best performance of few-shot in-context learning and fine-tuning. We can observe that the generated summaries of \Prompting{} are much better with few-shot inference/fine-tuning. However, we can also observe that both methods introduce extraneous information, potentially leading to hallucinations.

\end{document}